%% file: acl2020.tex
\PassOptionsToPackage{dvipsnames}{xcolor}
\documentclass[11pt,a4paper]{article}
\usepackage{hyperref}
\usepackage[hyperref]{acl2020}
\usepackage{xcolor}
\usepackage{times}
\usepackage{latexsym}

\usepackage{booktabs}
\usepackage{tabularx}
\usepackage{graphicx}
\usepackage[space]{grffile}

\usepackage{multirow}
\usepackage{makecell}
\usepackage{amsmath}
\usepackage[ruled,vlined]{algorithm2e}
\DeclareMathOperator*{\argmax}{arg\,max}

\usepackage{microtype}

\aclfinalcopy 
\def\aclpaperid{2725}

\title{Asking and Answering Questions to Evaluate the Factual Consistency of Summaries}

\author{Alex Wang\footnote{Work done while interning at Facebook.} \\
  New York University \\
  \texttt{alexwang@nyu.edu} \\\And
  Kyunghyun Cho \\
  Facebook AI \\
  New York University \\\And
  Mike Lewis \\
  Facebook AI
  \\}

\date{}

\begin{document}
\maketitle
\begin{abstract}
Practical applications of abstractive summarization models are limited by frequent factual inconsistencies with respect to their input.
Existing automatic evaluation metrics for summarization 
are largely insensitive to such errors.
We propose an automatic evaluation protocol called QAGS\footnote{Pronounced ``kags''.} that is designed to identify factual inconsistencies in a generated summary. 
QAGS is based on the intuition that if we ask questions about a summary and its source, we will receive similar answers if the summary is factually consistent with the source.
To evaluate QAGS, we collect human judgments of factual consistency on model-generated summaries for the CNN/DailyMail \citep{hermann2015teaching} and XSUM \citep{xsum-emnlp} summarization datasets.
QAGS has substantially higher correlations with these judgments than other automatic evaluation metrics.
Also, QAGS offers a natural form of interpretability: The answers and questions generated while computing QAGS indicate which tokens of a summary are inconsistent and why.
We believe QAGS is a promising tool in automatically generating usable and factually consistent text.
\end{abstract}

\section{Introduction}

Automatic summarization aims to produce summaries that are succinct, coherent, relevant, and --- crucially --- factually correct.
Recent progress in conditional text generation has led to models that can generate fluent, topical summaries \citep{lewis2019bart}. However, model-generated summaries frequently contain factual inconsistencies, limiting their applicability \citep{kryscinski2019neural}. 

The problem of factual inconsistency is due in part to the lack of automatic evaluation metrics that can detect such errors.
Standard metrics for evaluating generated text are predominantly based on counting $n$-grams, which weigh all $n$-grams equally and are insensitive to semantic errors.
This inadequacy leaves human evaluation as the primary method for evaluating the factual consistencies, which has been noted to be challenging even for humans \citep{daume2005bayesian,kryscinski2019evaluating}, in addition to being slow and costly.

We argue that evaluation metrics that are able to capture subtle semantic errors are required to build better models.
In this work, we introduce a general framework for evaluating conditional text generation that is designed to detect factual inconsistencies in generated text with respect to some input.
Our framework consists of three steps:
(1) Given a generated text, a question generation (QG) model generates a set of questions about the text. 
(2) We then use question answering (QA) models to answer these questions given both the input and the generated text. 
(3) A quality score is computed based on the similarity of corresponding answers.

This approach leverages recent progress in QA and QG to ask and answer human readable, on-topic questions \citep{devlin2019bert,song2019mass}. It only assumes access to a question answering dataset to train the QG and QA models, and is applicable to any modality where a QA model is available, e.g. text, images, or knowledge graphs.

We use this framework to develop QAGS (Question Answering and Generation for Summarization), a metric for evaluating the factual consistency of abstractive document summaries.
Compared to commonly used automatic metrics such as ROUGE \citep{lin2004rouge}, QAGS shows dramatically higher correlations with human judgements of factuality, for example achieving a Pearson correlation coefficient of 54.52 on the CNN/DailyMail summarization task, compared to 17.72 for ROUGE-2. 
QAGS also achieves new state-of-the-art results on evaluating the factuality of summaries, outperforming recently proposed NLI models for this task \citep{kryscinski2019evaluating}.

Finally, we analyse the robustness of QAGS through an ablation study. QAGS shows robustness to the quality of the underlying QG and QA models, the domain of the models, and the number of questions asked.
Even under the worst ablation settings, QAGS still has stronger correlation with human judgments than other automatic metrics.

Overall, we contribute the following:
(1) We introduce QAGS, an automatic model-based evaluation metric for measuring the factual consistency of model-generated text. 
(2) We collect a new set of human judgments of factual consistency of model-generated summaries for two summarization datasets. We demonstrate that QAGS correlates with these judgments significantly better than other automatic metrics.
(3) We show via ablations that QAGS is robust to a number of factors including underlying model quality and domain mismatch.
(4) We analyze the questions and answers produced in computing QAGS to illustrate which parts of summaries are inconsistent. 
(5) We will release models and code to compute QAGS.

\section{Background: Automatically Evaluating Machine Generated Text}\label{sec:background}

Standard approaches to evaluating generated text are primarily based on counting $n$-gram overlap.
These methods assume access to one or more reference texts, and score a generated summary based on the precision and recall of all reference $n$-grams in the generated summary.
We briefly describe the most common metrics in this family, and refer readers to \citet{liu2016not} for further discussion.

ROUGE \citep{lin2004rouge} was developed specifically for evaluating automatic summarization, and its variants are the \textit{de facto} standard for such.
The most common variant is ROUGE-$n$ (typically $n \in \{1, 2\}$), which computes the F1 score for all reference $n$-grams in the generated summary.
ROUGE-$L$, another commonly used variant, is the length of the longest common subsequence (possibly non-consecutive) between a summary and references.

BLEU \citep{papineni2002bleu} is closely related to ROUGE but was developed for machine translation. BLEU computes the precision of the reference $n$-grams in the generated summary.
METEOR \citep{lavie2007meteor} extends BLEU by using an alignment between the generated text and a reference, as well as using stemming and synonym replacement for more flexible $n$-gram matching.

We identify two key deficiencies when using these $n$-gram based evaluation metrics to detect factual inconsistencies in generated text.

First, these metrics require one or more reference texts to compare against.
Obtaining references can be expensive and challenging, and as such many text generation datasets contain only a single reference.
This problem is exacerbated with high-entropy generation tasks, such as summarization or dialogue, where there is a very large number of acceptable outputs.
In these settings, comparing against a single reference is woefully inadequate.

Second, given a reference to compare against, $n$-gram based approach weigh all portions of the text equally, even when only a small fraction of the $n$-grams carry most of the semantic content. Factual inconsistencies caused by minor changes may be drowned out by otherwise high $n$-gram overlap, making these metrics insensitive to these errors.
For example, the sentences ``I am writing my paper in Vancouver.'' and ``I am not writing my paper in Vancouver.'' share nearly all unigrams and bigrams despite having the opposite meaning.

\section{A Framework for Automatically Evaluating Factual Consistency}

\begin{figure*}
    \centering
    \includegraphics[width=\linewidth]{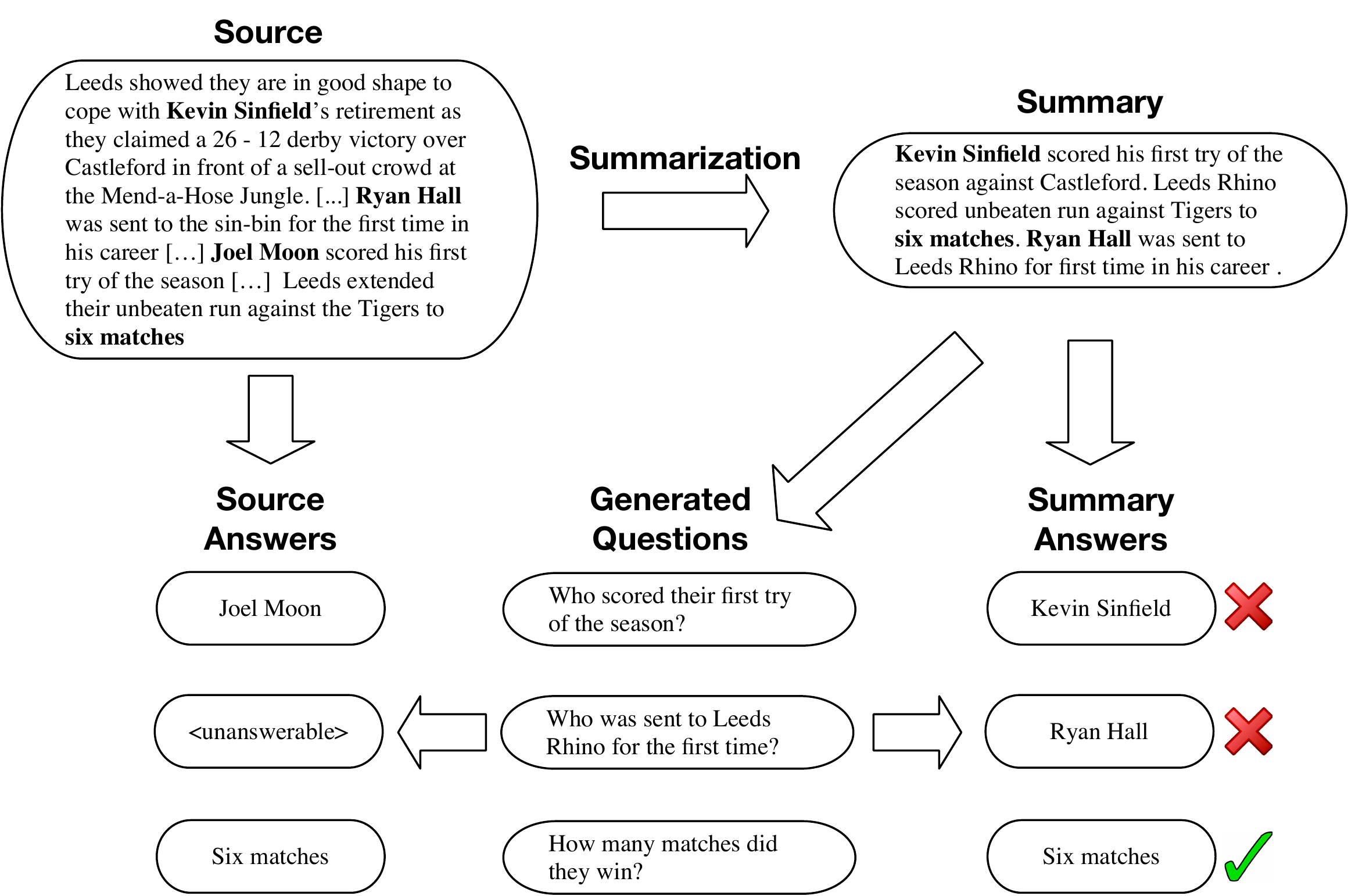}
    \caption{Overview of QAGS. A set of questions is generated based on the summary. The questions are then answered using both the source article and the summary. Corresponding answers are compared using a similarity function and averaged across questions to produce the final QAGS score.}
    \label{fig:figure}
\end{figure*}

We introduce a framework for automatically detecting factual inconsistencies in generated text while also addressing the deficiencies of current approaches.
Let $X$ and $Y$ be sequences of tokens coming from a vocabulary $V$ where $X$ is a source text and $Y$ is a summary of $X$. 
We define $p(Q|Y)$ as a distribution over all possible questions $Q$ given summary $Y$, and $p(A|Q, X)$ and $p(A|Q, Y)$ as distributions over all possible answers $A$ to a particular question $Q$ given either the source $X$ or the summary $Y$. We constrain the questions $Q$ and answers $A$ to also be sequences of tokens from $V$.
Then the factual consistency of the summary $Y$ is

\begin{equation}
\label{eq:qags}
    E_{Q \sim p(Q|Y)} \big[ D \big( p(A|Q, X),  p(A|Q, Y) \big) \big],
\end{equation}

where $D$ is some function measuring the similarity of the two answer distributions.
This expression is maximized when $Y$ contains a subset of the information in $X$ such that it produces the same answer for any question from $p(Q|Y)$.
This happens trivially when $Y=X$, e.g. we take $X$ as its own summary, but we usually have other desiderata of $Y$ such that this solution is undesirable.

This framework addresses the two issues with $n$-gram based approaches.
Instead of requiring a reference to compare against, our framework asks questions based on the generation itself, and compares answers with the provided source text.
Also, the use of questions focuses the metric on the semantically relevant parts of the generated text, rather than weighting all parts of the text equally. 


In practice, exactly computing the expectation in Equation~\ref{eq:qags} is intractable due to the large space of possible questions.
One potential workaround is to randomly sample questions from $p(Q|Y)$, but this suffers from high variance and requires many samples to obtain a good estimate.
Instead, we focus on producing highly probable questions, e.g. as produced by beam search, which may be biased in the limit, but will require fewer questions to estimate because of the higher quality of the questions.

\section{QAGS}

Using this framework requires specifying the question distribution $p(Q|Y)$, the answer distribution $p(A|Q, Y)$ (or $X$), and the answer similarity function $D$.
We apply this framework to summarization to develop QAGS and describe our instantiations of these components.

\paragraph{Question Generation} 
To instantiate $p(Q|Y)$, we draw on recent work on automatic question generation (QG), which models this distribution using neural seq2seq models \citep{du2017learning,krishna2019generating}.
We over-sample questions, and then filter out low quality questions as follows.

First, we train and generate from answer-conditional QG models: The model receives both the answer and the source article, and is trained to maximize the likelihood of the paired question.
At test time, we extract named entities and noun phrases as answers candidates using spaCy.\footnote{\url{https://spacy.io/api/entityrecognizer}}

Second, we filter out low-quality questions using a number of heuristics, such as duplicates and questions less than three tokens long.
We also found it useful to run the QA model (see next section) on all of the candidate questions, and filter out questions for which the QA model predicted no answer.

\paragraph{Question Answering}
We instantiate the answer distributions $p(A|Q,*)$ as extractive QA models, for simplicity.
We use extractive QA because we assume the facts are represented as text spans in the article and summary.
Future work should explore using abstractive QA models, which could match paraphrases of the same answer.

\paragraph{Answer Similarity} 
We use token-level F1 to compare answers, which is standard for extractive QA
and equivalent to defining $D$ as
$$F1(\argmax p(A|Q, X), \argmax p(A|Q, Y))$$

\paragraph{The QAGS Score}
Given these components, we obtain the QAGS score of a generation by (1) generating $K$ questions conditioned on the summary, (2) answering the questions using both the source article and the summary to get two sets of answers, (3) comparing corresponding answers using the answer similarity metric, and (4) averaging the answer similarity metric over all questions.
We depict this process in Figure~\ref{fig:figure}.

\section{Experiments}

\subsection{Human Evaluation}

We test whether QAGS accurately measures the factual consistency of a summary with respect to a source article by computing correlations with human judgments of factual consistency.

\paragraph{Datasets} We evaluate on two abstractive summarization datasets, CNN/Daily Mail \citep[CNNDM,][]{hermann2015teaching,nallapati-etal-2016-abstractive} and XSUM \citep{xsum-emnlp}.
Abstractive summarization is particularly interesting because factual consistency with the original text is crucial to usability, and a lack of such consistency has plagued abstractive neural summarization models \citep[][ i.a.]{cao2018faithful,falke2019ranking,kryscinski2019evaluating}.

CNN/DM is a standard dataset for summarization that consists of CNN and DailyMail articles.
Each reference summary consists of the concatenation of three editor-written, bullet point highlights.
For summaries, we use 235 test outputs from \citet{gehrmann2018bottom}.

XSUM was created by taking the first sentence of a news  article as the summary, and using the rest of the article as the source.
Consequently, XSUM summaries are significantly more abstractive than those of CNN/DM, and extractive summarization models perform poorly on this dataset.

We found that while the XSUM summaries are more abstractive, frequently there are facts (e.g. first names) in the summary that are not available in the ``article''.
This quirk made it especially difficult for humans and QAGS to tell when factual errors were being made by the summarization model.
To remedy this, for human evaluation and QAGS, we prepend the summary back to the ``article''.
We use a subset of 239 test outputs from BART fine-tuned on XSUM \citep{lewis2019bart}.

\input{tables/summary-corr.tex}

\paragraph{Annotation Protocol} 
We collect human judgments on Amazon Mechanical Turk\footnote{\url{https://www.mturk.com/}} via ParlAI \citep{miller2017parlai}.
We present summaries one sentence at a time, along with the entire article. 
For each summary sentence, the annotator makes a binary decision as to whether the sentence is factually consistent with the article.
Workers are instructed to mark non-grammatical sentences as not consistent, and copies of article sentences as consistent.
Workers are paid \$$1$ per full summary annotated.
See Appendix~\ref{ax:human} for further details.

We collect 3 annotations per summary.
To obtain a single ``correctness'' score per summary, we first take the majority vote for each sentence, then average the binary scores across summary sentences.

Inter-annotator agreement as measured by Krippendorff's $\alpha$ is 0.51 and 0.34 for CNN/DM and XSUM, respectively indicating ``moderate'' and ``fair'' agreement \citep{ageeva-etal-2015-evaluating}.
While not perfect, these agreement numbers are in-line with similar figures from previous work on summarization evaluation \citep{daume2005bayesian}.

\subsection{Experimental Details}

\paragraph{Question Generation}
We use \texttt{fairseq} \citep{ott2019fairseq} to fine-tune a pretrained BART language model on NewsQA \citep{trischler2017newsqa}, a dataset consisting of CNN articles and crowdsourced questions. 
For each summary, we use 10 answer candidates and generate questions using beam search with width 10, for a total of 100 question candidates.
After filtering, we use the $K = 20$ most probable questions.
If a summary has too few filtered questions, we randomly sample questions to reach the required number.
For details, see Appendix~\ref{ax:tuning}.

\paragraph{Question Answering}
We train QA models by fine-tuning BERT \citep{devlin2019bert} on SQuAD2.0 \citep{rajpurkar2018know}.
We use the \texttt{large-uncased} BERT variant via the \texttt{transformers} library \citep{wolf2019transformers}.

\paragraph{Baselines}
We compare against a number of automatic evaluation metrics: ROUGE \citep{lin2004rouge}, METEOR \citep{lavie2007meteor}, BLEU \citep{papineni2002bleu}, and BERTScore \citep{zhang2019bertscore}.
The latter uses BERT representations to compute an alignment between generation and reference tokens, and which is then used to compute a soft version of unigram F1. We use the \texttt{large-uncased} BERT variant.

\subsection{Results} 

We present results in Table \ref{tab:summary-correlations}. 
QAGS strongly outperforms other automatic evaluation metrics in terms of correlation with human judgments of factual consistency. BLEU and ROUGE perform comparably, and lower order $n$-gram metrics work better. 
BERTScore matches the best $n$-gram metrics on CNN/DM, but the worst overall on XSUM.

On CNN/DM, QAGS obtains nearly twice the correlation of the next best automatic metric (BLEU-1).
We speculate that this large increase is due to the sensitivity of the QA model to the sentence fusing behavior exhibited in many summarization models trained on CNN/DM \citep{lebanoff2019analyzing}.
When two sentences are fused to produce an incorrect summary statement, the QA model produces different answers than when using the source article versus when using the summary.

On XSUM, all metrics correlate worse with human judgments than on CNN/DM, which reflects the fact that XSUM is more abstractive.
QAGS still outperforms the next best automatic metric.

\subsection{Ablations}

A potential issue with model-based evaluation is that the quality of the evaluation metric may depend heavily on specific hyperparameter settings.
We explore whether this is true with QAGS by performing ablations on several factors.

\input{tables/ablation-qa-perf.tex}
\input{tables/ablation-qg-perf.tex}
\input{tables/ablation-n-qsts.tex}

\paragraph{Model Quality} We first consider the degree to which the quality of the underlying models impacts their evaluation capabilities.

For QA quality, we answer this question by training QA models of varying quality by fine-tuning different versions of BERT on SQuAD.
We present results in Table~\ref{tab:ablations-qa-perf}.
The QA models perform similarly despite substantially different performances on the SQuAD development set. 
Surprisingly, using the best QA model (\texttt{bert-large-wwm}) does not lead to the best correlations with human judgments.
On CNN/DM, \texttt{bert-large-wwm} slightly underperforms \texttt{bert-base} and \texttt{bert-large}.
On XSUM, \texttt{bert-base} slightly outperforms the other two BERT variants.
These results indicate that QAGS is fairly robust to the quality of the underlying QA model, though we note that BERT is a strong QA baseline, and using weaker QA models might lead to larger performance dropoffs.

To ablate QG quality, we use models with increasing perplexity on the NewsQA development set.
Results in Table~\ref{tab:ablations-qg-perf} show that QAGS is  robust to the QG model quality, with some decrease in correlation with human judgments as perplexity increases on CNN/DM, and no clear trend on XSUM.
Even the weakest QG model still significantly outperforms all other automatic metrics in Table~\ref{tab:summary-correlations}.

\paragraph{Domain Effects}
Our approach relies on having a labeled dataset to train QG and QA models.
However, for relatively niche domains, such a labeled QA/QG dataset may not exist.
Instead, we may need to resort to using models trained on out-of-domain data, leading to domain shift effects that negatively impact the quality of the QAGS scores.
We simulate this setting by fine-tuning the QG model on SQuAD, which is of similar size to NewsQA but drawn from Wikipedia articles rather than CNN articles, which exactly matches the genre of the summarization datasets.

Evaluating with this QG model, we get correlations of 51.53 and 15.28 with human judgments on CNN/DM and XSUM respectively, versus 54.53 and 17.49 when using the NewsQA-tuned QG model.
The drop in performance indicates a negative domain shift effect.
However using the SQuAD-tuned QG model still substantially outperforms all other automatic metrics, again pointing to the robustness of QAGS.

\paragraph{Number of Questions}
Next, we investigate the correlation with human judgments when varying the number of questions used.
Results in Table~\ref{tab:ablations-n-qsts} show that increasing the number of questions used improves correlations with human judgments.
We observe a large increase when moving from 10 to 20 questions, and a smaller increase from 20 to 50 questions, indicating decreasing marginal benefit moving beyond 50 questions.
With just 5 questions, QAGS still substantially outperforms other automatic metrics, indicating its robustness.

\paragraph{Answer Similarity Metric}
Finally, we consider using exact match as an alternative answer similarity metric.
Exact match is another common evaluation metric for extractive QA, and is more restrictive than F1.
When using EM, we obtain Pearson correlations with human judgments of 45.97 and 18.10 on CNN/DM and XSUM, as opposed to 54.53 and 17.49 when using F1.

\section{Re-ranking with QAGS}

%
%

Several works explore the use of natural language inference (NLI) models to detect factual consistency in generated text \citep{welleck2018dialogue,falke2019ranking}.
We compare against these methods by evaluating on the sentence ranking experiment from \citet{falke2019ranking}.
The experiment uses 373 triplets of source sentences from CNN/DM and two summary sentences generated from the model from \citet{chen2018fast}.
One summary sentence is factually consistent with the source sentence, and the other is inconsistent.
A metric (or model) is evaluated based on how often it ranks the consistent sentence higher than the inconsistent sentence. 

\input{tables/falke-sent-rank.tex}

We present the results in Table~\ref{tab:falke-sent-rank}.
Results using two NLI models fine-tuned on MultiNLI \citep{williams2018broad}, BERT NLI and ESIM \citep{chen2017enhanced}, are from \citet{falke2019ranking}.
FactCC \citep{kryscinski2019evaluating} is an NLI-based fact-checking model that is trained on a dataset tailor made for detecting factual inconsistencies in generated text.
QAGS outperforms these methods, while requiring no special supervision for this task.

\input{tables/qags-examples.tex}

\section{Qualitative Analysis}\label{sec:analysis}


\paragraph{Interpreting QAGS}  
The questions and answers produced in computing QAGS are directly interpretable, and highlight errors in summaries.
We present examples of articles, summaries, and the QAGS questions and answers in Table~\ref{tab:examples}.

On the first example (Table~\ref{tab:examples}, top), QAGS detects several factual inconsistencies in the generated summary: The summary mistakes the first name of the attacker, the location of the attack, and the weapons used. Because the QG model focuses on these details, QAGS is able to correctly penalize the summary for its hallucinations. 
Because the answer candidates used are mostly named entities and noun phrases, QAGS is particularly effective at detecting errors of this kind.
Using more diverse answer candidates may broaden the set of inconsistencies that QAGS is able to detect.

The second example (Table~\ref{tab:examples}, bottom), illustrates failure modes of QAGS. For example, the QA model incorrectly marks question 2 as unanswerable. On question 4, both answers produced are correct, but because they have no common tokens, they are marked inconsistent by QAGS.

\paragraph{Error Analysis} 
The interpretability of QAGS allows for error analysis on the metric.
We manually annotate 400 triplets of generated questions, article answers, and summary answers that are produced in computing QAGS on the XSUM summaries, and label them by the quality of the generated questions, predicted answers, and answer similarity scores.

Among the generated questions, 8.75\% are nonsensical, while 3.00\% are well-formed but unanswerable using the generated summary they were conditioned upon.
These figures indicate that the vast majority of questions are understandable and on-topic. 
We frequently observe multiple questions with slightly different wordings, which is likely due to the low number of answer candidates in XSUM summaries (which are one sentence long) and due to beam search.
8.25\% of questions are well-formed but unanswerable using the source, which is usually due to a hallucinated fact in the summary that the QG model turns into a question.

Among predicted answers, 1.75\% of questions are potentially answerable using the summary, but are  incorrectly answered.
This percentage increases to 32.50\% for the article, which indicates that the transfer ability of the QA model is lacking.
In a small number of cases, we found that while a question had a single answer in the summary, it could have multiple answers in the article.

Finally, for 8.00\% of the examples, the question is answered correctly using both the article and summary, but the answers have high lexical variation such that F1 score fails to detect their similarity. 
While this happens in a relatively small number of cases, exploring similarity metrics other than $n$-gram based approaches could be useful.


\paragraph{Limitations}

We emphasize that QAGS and our overall framework are specifically designed to detect factual inconsistencies in generated summaries relative to the source article. 
QAGS does not measure other desirable properties of generated text, including fluency, readability, or factual recall. 
We therefore recommend using QAGS in conjunction with complementary evaluation metrics.

The choices of QG and QA models in QAGS are particular to abstractive summarization and may require adaptation to be used for other conditional text generation tasks.
For example, we expect that extractive summarization models may obtain nearly perfect QAGS scores because facts and statements are directly copied from the source article.

\section{Related Work}

Automatic summarization and its evaluation are long-standing lines of work in NLP, dating at least as far back as the Document Understanding Conferences \citep{chali2004summarization}.
The primary evaluation metric then and now is ROUGE \citep{lin2004rouge},
though much work has demonstrated the limited ability of ROUGE and its relatives to evaluate summaries \citep[][i.a.]{dorr2004extrinsic,liu2009exploring,kedzie2018content}.
Other metrics have focused on specific aspects of summarization quality, including content selection \citep{nenkova2004evaluating}, relevance prediction \citep{daume2005bayesian}, and many more.

There has been a recent resurgence of work leveraging NLU models for evaluating the factuality of generated text.
\citet{goodrich2019assessing} use information extraction  models to measure factual overlap, but facts are restricted to pre-defined schemas.
\citet{falke2019ranking} investigate the use of NLI models to evaluate the factual correctness of CNN/DM summaries, and conclude that current NLI models are too brittle to be reliably used in this manner.
\citet{kryscinski2019evaluating} train a NLI-based fact-checking model by building a dataset of factual inconsistencies based on noise heuristic.
Our QA approach allows a finer-grained analysis, because NLI operates on complete sentences, whereas QAGS can ask many questions about the same sentence.

Most relatedly, \citet{eyal2019question} and \citet{scialom2019answers} use QA models to evaluate summarization.
We diverge from these works in two important ways.
First, both works use Cloze-style questions, which are generated by masking entities in either the source document or the reference summary.
We instead generate the questions with a model, allowing a much greater range of questions.
Second, we produce questions conditioned on the generated summary, rather than the reference summary or source article.
Producing questions from the generated summary is more appropriate for verifying the accuracy of the text, whereas using the reference or source measures content selection.


\section{Conclusion}

We introduce a framework for automatically detecting factual inconsistencies in conditionally generated texts and use this framework to develop QAGS, a metric for measuring inconsistencies in abstractive summarization.
QAGS correlates with human judgments of factuality significantly better than standard automatic evaluation metrics for summarization, and outperforms related NLI-based approaches to factual consistency checking. 
QAGS is naturally interpretable: The questions and answers produced in computing QAGS indicate which tokens in a generated summary are inconsistent and why.
Error analysis shows that future work should explore improved QA models. 
Our approach can also be applied to diverse modalities, such as translation and image captioning.
Overall, we believe QAGS is useful in quantifying and incentivizing factually consistent text generation.

%

%

\bibliography{acl2020}
\bibliographystyle{natbib}

\clearpage
\appendix

\section{Human Evaluation Task Design}\label{ax:human}

\begin{figure*}[h]
    \centering
    \includegraphics[width=\linewidth]{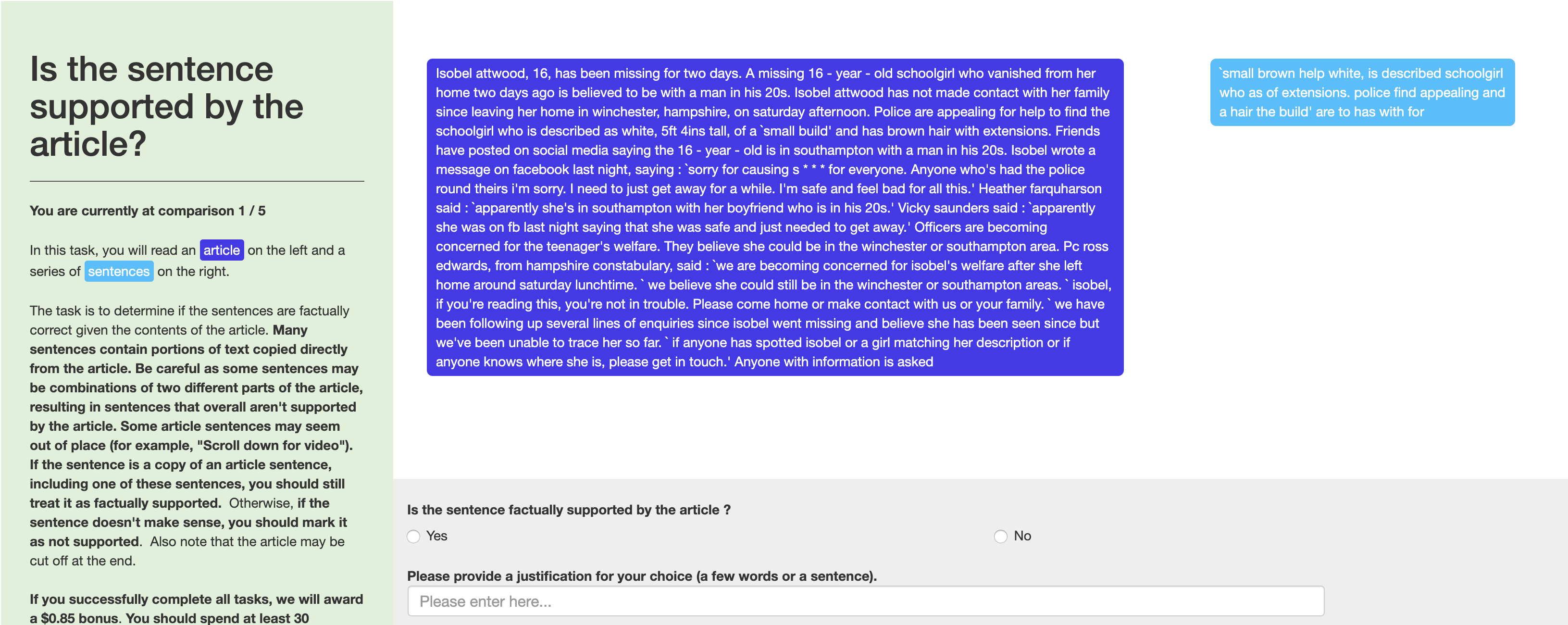}
    \caption{Annotation interface and instructions for CNN/DM factual consistency task.}
    \label{fig:mturk_cnndm}
\end{figure*}

\begin{figure*}[h]
    \centering
    \includegraphics[width=\linewidth]{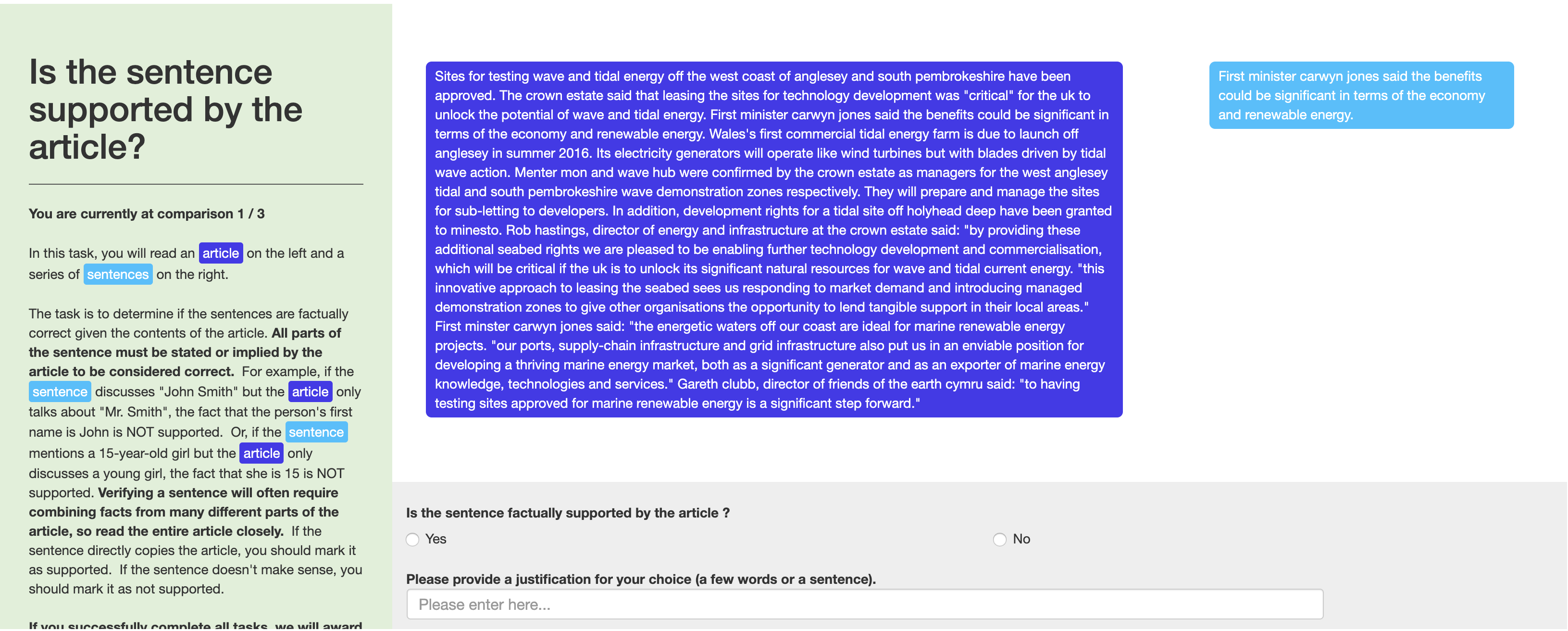}
    \caption{Annotation interface and instructions for XSUM factual consistency task.}
    \label{fig:mturk_xsum}
\end{figure*}

We restrict our pool of workers to US-based workers.
Workeres are required to have at least 1000 approved HITs with an acceptance rate of at least 98\%.

The base reward for our task is \$0.15.
For each summary, we include automatic quality checks including
\begin{itemize}
    \item Time checks: workers who complete the task under 30s fail the check
    \item Attention checks: we include exact copies of article sentences and corrupted mixtures of two article sentences as positive and negative control task. If a worker fails to answer both of these examples correctly, they fail the check
    \item Explanation checks: For each sentence in the summary, the worker is required to provide a short explanation of their decision
\end{itemize}
If a worker passes all checks, they are awarded a \$0.85 bonus, totalling \$1.00 per correct annotation. 
According to \url{turkerview.com}, workers of our HIT are paid well in excess of \$15.00 on average.

We show our annotation interfaces for the annotation task for CNN/DM and XSUM respectively in Figures~\ref{fig:mturk_cnndm} and \ref{fig:mturk_xsum}.
We use slightly different instructions to accommodate for the quirks of each dataset. For XSUM, we prepend the reference ``summary'' back onto the source article, as without it, workers were struggling to identify factual inconsistencies.

\section{Model and Generation Details}\label{ax:tuning}

\paragraph{Question Generation}
We fine-tune BART for question generation using the same tuning hyperparameters as the original work.
We optimize label smoothed cross entropy with smoothing parameter 0.1 \citep{pereyra2017regularizing} and a peak learning rate of 2e-5.
We optimize for 100k steps with 5k warmup steps, and use the model with the best perplexity on the development set.

To turn NewsQA into an answer conditional QG dataset, we concatenate the answer to the source article with a special marker token in between. We then concatenate another special marker token and the question.
At test time, we get 10 named entities and noun phrases as answer candidates using the \texttt{en-web-sm} spaCy model. We downsampling if there are more than 10 and randomly duplicating some answers if there are more than 10.
The model predicts the question after seeing an answer and the article.

During decoding, we use beam search with beam size 10, length penalty 1.0, and trigram repetition blocking.
We experimented with top-$k$ \citep{holtzman2019curious} and top-$p$ \citep{fan2018hierarchical}, but the outputted questions, while diverse, were quite noisy.
Generations have minimum length 8 and max length 60.

To filter the questions, we first use simple heuristics, including removing
\begin{itemize}
    \item everything after the first question mark in a question
    \item exact duplicates
    \item questions shorter than three tokens long
\end{itemize}
For the remaining questions, we use our QA model to answer each question and we remove questions for which the QA model deems unanswerable.
We then take the top 20 most probable questions, random sampling some of the filtered questions if there were too few.

\paragraph{Question Answering}
We fine-tune BERT for question answering following the original work.
We optimize using AdamW \citep{loshchilov2018decoupled} with initial learning rate 5e-5.
We train for 3 epochs, with a warmup ratio of 0.1.
We use the model with the best development set performance.

We use SQuAD2.0 because we found the unanswerable questions useful for filtering out questions and questions based on hallucinated facts in the summary should be unanswerable using the source article. Similar to the QG setting, we append the question and answer to the source article with intervening special marker tokens.

\end{document}

%% file: tables/summary-corr.tex
\begin{table}[t]
    \centering
    \begin{tabular}{lcc}
        \toprule
        Metric & CNN/DM & XSUM \\
        \midrule
        
        ROUGE-1 & 28.74 & 13.22  \\
        ROUGE-2 & 17.72 & 8.95   \\
        ROUGE-L & 24.09 & 8.86   \\
        METEOR  & 26.65 & 10.03  \\
        BLEU-1  & 29.68 & 11.76  \\
        BLEU-2  & 25.65 & 11.68  \\
        BLEU-3  & 23.96 & 8.41 \\
        BLEU-4  & 21.45 & 5.64 \\
        BERTScore & 27.63 & 2.51 \\
        QAGS & \textbf{54.53} & \textbf{17.49} \\       
        \bottomrule
    \end{tabular}
    \caption{Summary-level Pearson correlation coefficients between various automatic metrics and human judgments of correctness for summarization datasets. 
    QAGS obtains substantially higher correlations than all other automatic metrics.}
    \label{tab:summary-correlations}
\end{table}

%% file: tables/ablation-qa-perf.tex
\begin{table}[t]
    \small
    \centering
    \begin{tabular}{lccc}
        \toprule
        \multirow{2}{*}{QA model} & SQuAD & CNN/DM & XSUM \\
         & (F1) & (Pear.) & (Pear.) \\
        \midrule
        \texttt{bert-base}      & 75.95 & 55.20 & 20.71 \\
        \texttt{bert-large}     & 81.57 & 54.53 & 17.49 \\
        \texttt{bert-large-wwm} & 84.36 & 51.36 & 18.07 \\
        \bottomrule
    \end{tabular}
    
    \caption{Pearson correlations between human judgments of factual consistency and QAGS using QA models of different qualities, as measured by performance on the SQuAD2.0 development set (F1). The correlations are stable across QA model quality.}
    \label{tab:ablations-qa-perf}
\end{table}

%% file: tables/ablation-qg-perf.tex
\begin{table}[t]
    \centering
    \begin{tabular}{ccc}
        \toprule
        NewsQA & CNN/DM & XSUM \\
        (ppl.) & (Pear.) & (Pear.) \\
        \midrule
        5.48  & 54.53 & 17.49 \\
        9.50  & 50.09 & 19.93 \\ 
        18.56 & 47.92 & 16.38 \\ 
        \bottomrule
    \end{tabular}
    
    \caption{Pearson correlations between human judgments of factual consistency and QAGS with QG models of varying quality, as measured by perplexity on the NewsQA development set. We see some decrease in correlation on CNN/DM as QG perplexity increases, though we do not see a similar trend for XSUM.}
    \label{tab:ablations-qg-perf}
\end{table}

%% file: tables/ablation-n-qsts.tex
\begin{table}[t]
    \centering
    \begin{tabular}{ccc}
        \toprule
        \# Questions & CNN/DM & XSUM \\
        \midrule
        5 &  41.61 & 15.63 \\
        10 & 41.17 & 15.49 \\
        20 & 54.53 & 17.49 \\
        50 & 57.94 & 17.74 \\
        \bottomrule
    \end{tabular}
    \caption{Pearson correlation coefficients between QAGS scores with varying number of questions and human judgments of correctness for summarization datasets. The correlation increases with the number of questions used, but with decreasing marginal benefit.}
    \label{tab:ablations-n-qsts}
\end{table}

%% file: tables/falke-sent-rank.tex
\begin{table}[t]
    \centering
    \begin{tabular}{lc}
        \toprule
        Model/Metric & \% Correct ($\uparrow$) \\
        \midrule
        Random & 50.0\% \\
        BERT NLI & 64.1\% \\
        ESIM & 67.6\% \\
        FactCC & 70.0\% \\
        QAGS & \textbf{72.1\%} \\
        \bottomrule
    \end{tabular}
    \caption{Results on the sentence ranking task from \citet{falke2019ranking}. Results using BERT NLI and ESIM are from \citet{falke2019ranking}; FactCC is from \citet{kryscinski2019evaluating}. QAGS outperforms previous work.}
    \label{tab:falke-sent-rank}
\end{table}

%% file: tables/qags-examples.tex
\begin{table*}[t]
    \centering
    \begin{tabularx}{\linewidth}{X}
        \toprule

        \textbf{Article:} On Friday, 28-year-old Usman Khan stabbed reportedly several people at Fishmongers' Hall in London with a large knife, then fled up London Bridge. Members of the public confronted him; one man sprayed Khan with a fire extinguisher, others struck him with their fists and took his knife, and another, a Polish chef named Łukasz, harried him with a five-foot narwhal tusk. [\dots] \\
        \textbf{Summary :} On Friday afternoon , a man named Faisal Khan entered a Cambridge University building and started attacking people with a knife and a fire extinguisher . \\ 
        
        
        \textbf{Question 1:} What did the attacker have ? \\
        \textbf{Article answer:} {a large knife} \quad \textbf{Summary answer:} {a knife and a fire extinguisher} \\
        
        \textbf{Question 2:} When did the attack take place ? \\
        \textbf{Article answer:} {Friday} \quad \textbf{Summary answer:} {Friday afternoon} \\
        
         \textbf{Question 3:} What is the attacker's name ? \\
        \textbf{Article answer:} {Usman Khan} \quad \textbf{Summary answer:} {Faisal Khan} \\       
        
        \textbf{Question 4:} Where did the attack take place ? \\
        \textbf{Article answer:} {Fishmongers' Hall} \quad \textbf{Summary answer:} {Cambridge University building} \\

        \midrule

        \textbf{Article: } In findings published on Wednesday in the journal PLOS ONE, an international team of scientists report ancient Egyptians captured sacred ibises (Threskiornis aethiopicus) from the wild for use in ritual sacrifice rather than domesticating the birds. [\dots] The team collected DNA samples from mummified birds collected from six separate catacombs including sites at Abydos, Saqqara, and Tuna el-Gebel with permission from the Egyptian Ministry of State for Antiquity, and several museums offered to send tissue samples from the mummified ibises in their collections. [\dots] \\
        \textbf{Summary :} Archaeologists have used DNA samples from ancient ibis birds to determine whether the birds were domesticated or sacrificed in ancient Egypt \\ 
        
        \textbf{Question 1:} Archaeologists have used what to determine whether the birds were domesticated ? \\
        \textbf{Article Answer}: hatchery structures \quad
        \textbf{Summary Answer}: DNA samples \\

        
        \textbf{Question 2:} Who used DNA samples to determine whether the birds were domesticated ? \\
        \textbf{Article Answer:} [NO ANSWER] \quad
        \textbf{Summary Answer:} Archaeologists \\
        
        \textbf{Question 3:} What are archeologists using to determine whether the birds were domesticated ? \\
        \textbf{Article Answer:} DNA samples \quad
        \textbf{Summary Answer:} DNA samples \\
        
        \textbf{Question 4:} Where were the birds found? \\
        \textbf{Article Answer:} six separate catacombs \quad
        \textbf{Summary Answer:} ancient Egypt \\
        
        \bottomrule
    \end{tabularx}
    \caption{Example questions and answers generated when computing QAGS. The questions are overwhelmingly fluent and relevant. The answers indicate which tokens in the summary are factually consistent or inconsistent.}
    \label{tab:examples}
\end{table*}

%% file: acl2020.bbl
\begin{thebibliography}{42}
\expandafter\ifx\csname natexlab\endcsname\relax\def\natexlab#1{#1}\fi

\bibitem[{Ageeva et~al.(2015)Ageeva, Forcada, Tyers, and
  P{\'e}rez-Ortiz}]{ageeva-etal-2015-evaluating}
Ekaterina Ageeva, Mikel~L. Forcada, Francis~M. Tyers, and Juan~Antonio
  P{\'e}rez-Ortiz. 2015.
\newblock \href {https://www.aclweb.org/anthology/W15-4918} {Evaluating machine
  translation for assimilation via a gap-filling task}.
\newblock In \emph{Proceedings of the 18th Annual Conference of the {E}uropean
  Association for Machine Translation}, pages 137--144, Antalya, Turkey.

\bibitem[{Cao et~al.(2018)Cao, Wei, Li, and Li}]{cao2018faithful}
Ziqiang Cao, Furu Wei, Wenjie Li, and Sujian Li. 2018.
\newblock Faithful to the original: Fact aware neural abstractive
  summarization.
\newblock In \emph{Thirty-Second AAAI Conference on Artificial Intelligence}.

\bibitem[{Chali and Kolla(2004)}]{chali2004summarization}
Yllias Chali and Maheedhar Kolla. 2004.
\newblock Summarization techniques at duc 2004.
\newblock In \emph{In Proceedings of the Document Understanding Conference}.
  Citeseer.

\bibitem[{Chen et~al.(2017)Chen, Zhu, Ling, Wei, Jiang, and
  Inkpen}]{chen2017enhanced}
Qian Chen, Xiaodan Zhu, Zhen-Hua Ling, Si~Wei, Hui Jiang, and Diana Inkpen.
  2017.
\newblock Enhanced lstm for natural language inference.
\newblock In \emph{Proceedings of the 55th Annual Meeting of the Association
  for Computational Linguistics (Volume 1: Long Papers)}, pages 1657--1668.

\bibitem[{Chen and Bansal(2018)}]{chen2018fast}
Yen-Chun Chen and Mohit Bansal. 2018.
\newblock Fast abstractive summarization with reinforce-selected sentence
  rewriting.
\newblock In \emph{Proceedings of the 56th Annual Meeting of the Association
  for Computational Linguistics (Volume 1: Long Papers)}, pages 675--686.

\bibitem[{Daume~III and Marcu(2005)}]{daume2005bayesian}
Hal Daume~III and Daniel Marcu. 2005.
\newblock Bayesian summarization at duc and a suggestion for extrinsic
  evaluation.
\newblock In \emph{Proceedings of the Document Understanding Conference,
  DUC-2005, Vancouver, USA}.

\bibitem[{Devlin et~al.(2019)Devlin, Chang, Lee, and
  Toutanova}]{devlin2019bert}
Jacob Devlin, Ming-Wei Chang, Kenton Lee, and Kristina Toutanova. 2019.
\newblock Bert: Pre-training of deep bidirectional transformers for language
  understanding.
\newblock In \emph{Proceedings of the 2019 Conference of the North American
  Chapter of the Association for Computational Linguistics: Human Language
  Technologies, Volume 1 (Long and Short Papers)}, pages 4171--4186.

\bibitem[{Dorr et~al.(2004)Dorr, Monz, Oard, Zajic, and
  Schwartz}]{dorr2004extrinsic}
Bonnie Dorr, Christof Monz, Douglas Oard, David Zajic, and Richard Schwartz.
  2004.
\newblock Extrinsic evaluation of automatic metrics for summarization.
\newblock Technical report, MARYLAND UNIV COLLEGE PARK INST FOR ADVANCED
  COMPUTER STUDIES.

\bibitem[{Du et~al.(2017)Du, Shao, and Cardie}]{du2017learning}
Xinya Du, Junru Shao, and Claire Cardie. 2017.
\newblock Learning to ask: Neural question generation for reading
  comprehension.
\newblock In \emph{Proceedings of the 55th Annual Meeting of the Association
  for Computational Linguistics (Volume 1: Long Papers)}, pages 1342--1352.

\bibitem[{Eyal et~al.(2019)Eyal, Baumel, and Elhadad}]{eyal2019question}
Matan Eyal, Tal Baumel, and Michael Elhadad. 2019.
\newblock Question answering as an automatic evaluation metric for news article
  summarization.
\newblock In \emph{Proceedings of the 2019 Conference of the North American
  Chapter of the Association for Computational Linguistics: Human Language
  Technologies, Volume 1 (Long and Short Papers)}, pages 3938--3948.

\bibitem[{Falke et~al.(2019)Falke, Ribeiro, Utama, Dagan, and
  Gurevych}]{falke2019ranking}
Tobias Falke, Leonardo~FR Ribeiro, Prasetya~Ajie Utama, Ido Dagan, and Iryna
  Gurevych. 2019.
\newblock Ranking generated summaries by correctness: An interesting but
  challenging application for natural language inference.
\newblock In \emph{Proceedings of the 57th Conference of the Association for
  Computational Linguistics}, pages 2214--2220.

\bibitem[{Fan et~al.(2018)Fan, Lewis, and Dauphin}]{fan2018hierarchical}
Angela Fan, Mike Lewis, and Yann Dauphin. 2018.
\newblock Hierarchical neural story generation.
\newblock In \emph{Proceedings of the 56th Annual Meeting of the Association
  for Computational Linguistics (Volume 1: Long Papers)}, pages 889--898.

\bibitem[{Gehrmann et~al.(2018)Gehrmann, Deng, and Rush}]{gehrmann2018bottom}
Sebastian Gehrmann, Yuntian Deng, and Alexander Rush. 2018.
\newblock Bottom-up abstractive summarization.
\newblock In \emph{Proceedings of the 2018 Conference on Empirical Methods in
  Natural Language Processing}, pages 4098--4109.

\bibitem[{Goodrich et~al.(2019)Goodrich, Rao, Liu, and
  Saleh}]{goodrich2019assessing}
Ben Goodrich, Vinay Rao, Peter~J. Liu, and Mohammad Saleh. 2019.
\newblock \href {https://doi.org/10.1145/3292500.3330955} {Assessing the
  factual accuracy of generated text}.
\newblock In \emph{Proceedings of the 25th ACM SIGKDD International Conference
  on Knowledge Discovery \& Data Mining}, KDD '19, pages 166--175, New York,
  NY, USA. ACM.

\bibitem[{Hermann et~al.(2015)Hermann, Kocisky, Grefenstette, Espeholt, Kay,
  Suleyman, and Blunsom}]{hermann2015teaching}
Karl~Moritz Hermann, Tomas Kocisky, Edward Grefenstette, Lasse Espeholt, Will
  Kay, Mustafa Suleyman, and Phil Blunsom. 2015.
\newblock Teaching machines to read and comprehend.
\newblock In \emph{Advances in neural information processing systems}, pages
  1693--1701.

\bibitem[{Holtzman et~al.(2019)Holtzman, Buys, Forbes, and
  Choi}]{holtzman2019curious}
Ari Holtzman, Jan Buys, Maxwell Forbes, and Yejin Choi. 2019.
\newblock The curious case of neural text degeneration.
\newblock \emph{arXiv preprint arXiv:1904.09751}.

\bibitem[{Kedzie et~al.(2018)Kedzie, McKeown, and
  Daume~III}]{kedzie2018content}
Chris Kedzie, Kathleen McKeown, and Hal Daume~III. 2018.
\newblock Content selection in deep learning models of summarization.
\newblock In \emph{Proceedings of the 2018 Conference on Empirical Methods in
  Natural Language Processing}, pages 1818--1828.

\bibitem[{Krishna and Iyyer(2019)}]{krishna2019generating}
Kalpesh Krishna and Mohit Iyyer. 2019.
\newblock \href {https://doi.org/10.18653/v1/P19-1224} {Generating
  question-answer hierarchies}.
\newblock In \emph{Proceedings of the 57th Annual Meeting of the Association
  for Computational Linguistics}, pages 2321--2334, Florence, Italy.
  Association for Computational Linguistics.

\bibitem[{Kryscinski et~al.(2019{\natexlab{a}})Kryscinski, Keskar, McCann,
  Xiong, and Socher}]{kryscinski2019neural}
Wojciech Kryscinski, Nitish~Shirish Keskar, Bryan McCann, Caiming Xiong, and
  Richard Socher. 2019{\natexlab{a}}.
\newblock Neural text summarization: A critical evaluation.
\newblock In \emph{Proceedings of the 2019 Conference on Empirical Methods in
  Natural Language Processing, Volume 1 (Long and Short Papers)}.

\bibitem[{Kryscinski et~al.(2019{\natexlab{b}})Kryscinski, McCann, Xiong, and
  Socher}]{kryscinski2019evaluating}
Wojciech Kryscinski, Bryan McCann, Caiming Xiong, and Richard Socher.
  2019{\natexlab{b}}.
\newblock Evaluating the factual consistency of abstractive text summarization.

\bibitem[{Lavie and Agarwal(2007)}]{lavie2007meteor}
Alon Lavie and Abhaya Agarwal. 2007.
\newblock Meteor: An automatic metric for mt evaluation with high levels of
  correlation with human judgments.
\newblock In \emph{Proceedings of the Second Workshop on Statistical Machine
  Translation}, pages 228--231. Association for Computational Linguistics.

\bibitem[{Lebanoff et~al.(2019)Lebanoff, Muchovej, Dernoncourt, Kim, Kim,
  Chang, and Liu}]{lebanoff2019analyzing}
Logan Lebanoff, John Muchovej, Franck Dernoncourt, Doo~Soon Kim, Seokhwan Kim,
  Walter Chang, and Fei Liu. 2019.
\newblock Analyzing sentence fusion in abstractive summarization.
\newblock In \emph{Proceedings of the 2nd Workshop on New Frontiers in
  Summarization}, pages 104--110.

\bibitem[{Lewis et~al.(2019)Lewis, Liu, Goyal, Ghazvininejad, Mohamed, Levy,
  Stoyanov, and Zettlemoyer}]{lewis2019bart}
Mike Lewis, Yinhan Liu, Naman Goyal, Marjan Ghazvininejad, Abdelrahman Mohamed,
  Omer Levy, Ves Stoyanov, and Luke Zettlemoyer. 2019.
\newblock {BART}: Denoising sequence-to-sequence pre-training for natural
  language generation, translation, and comprehension.
\newblock \emph{arXiv preprint 1910.13461}.

\bibitem[{Lin(2004)}]{lin2004rouge}
Chin-Yew Lin. 2004.
\newblock Rouge: A package for automatic evaluation of summaries.
\newblock In \emph{Text summarization branches out}, pages 74--81.

\bibitem[{Liu et~al.(2016)Liu, Lowe, Serban, Noseworthy, Charlin, and
  Pineau}]{liu2016not}
Chia-Wei Liu, Ryan Lowe, Iulian Serban, Mike Noseworthy, Laurent Charlin, and
  Joelle Pineau. 2016.
\newblock How not to evaluate your dialogue system: An empirical study of
  unsupervised evaluation metrics for dialogue response generation.
\newblock In \emph{Proceedings of the 2016 Conference on Empirical Methods in
  Natural Language Processing}, pages 2122--2132.

\bibitem[{Liu and Liu(2009)}]{liu2009exploring}
Feifan Liu and Yang Liu. 2009.
\newblock Exploring correlation between rouge and human evaluation on meeting
  summaries.
\newblock \emph{IEEE Transactions on Audio, Speech, and Language Processing},
  18(1):187--196.

\bibitem[{Loshchilov and Hutter(2018)}]{loshchilov2018decoupled}
Ilya Loshchilov and Frank Hutter. 2018.
\newblock Decoupled weight decay regularization.

\bibitem[{Miller et~al.(2017)Miller, Feng, Batra, Bordes, Fisch, Lu, Parikh,
  and Weston}]{miller2017parlai}
Alexander Miller, Will Feng, Dhruv Batra, Antoine Bordes, Adam Fisch, Jiasen
  Lu, Devi Parikh, and Jason Weston. 2017.
\newblock Parlai: A dialog research software platform.
\newblock In \emph{Proceedings of the 2017 Conference on Empirical Methods in
  Natural Language Processing: System Demonstrations}, pages 79--84.

\bibitem[{Nallapati et~al.(2016)Nallapati, Zhou, dos Santos,
  GuÌ‡l{\c{c}}ehre, and Xiang}]{nallapati-etal-2016-abstractive}
Ramesh Nallapati, Bowen Zhou, Cicero dos Santos, {\c{C}}a{\u{g}}lar
  GuÌ‡l{\c{c}}ehre, and Bing Xiang. 2016.
\newblock \href {https://doi.org/10.18653/v1/K16-1028} {Abstractive text
  summarization using sequence-to-sequence {RNN}s and beyond}.
\newblock In \emph{Proceedings of The 20th {SIGNLL} Conference on Computational
  Natural Language Learning}, pages 280--290, Berlin, Germany. Association for
  Computational Linguistics.

\bibitem[{Narayan et~al.(2018)Narayan, Cohen, and Lapata}]{xsum-emnlp}
Shashi Narayan, Shay~B. Cohen, and Mirella Lapata. 2018.
\newblock Don't give me the details, just the summary! {T}opic-aware
  convolutional neural networks for extreme summarization.
\newblock In \emph{Proceedings of the 2018 Conference on Empirical Methods in
  Natural Language Processing}, Brussels, Belgium.

\bibitem[{Nenkova and Passonneau(2004)}]{nenkova2004evaluating}
Ani Nenkova and Rebecca Passonneau. 2004.
\newblock Evaluating content selection in summarization: The pyramid method.
\newblock In \emph{Proceedings of the human language technology conference of
  the north american chapter of the association for computational linguistics:
  Hlt-naacl 2004}, pages 145--152.

\bibitem[{Ott et~al.(2019)Ott, Edunov, Baevski, Fan, Gross, Ng, Grangier, and
  Auli}]{ott2019fairseq}
Myle Ott, Sergey Edunov, Alexei Baevski, Angela Fan, Sam Gross, Nathan Ng,
  David Grangier, and Michael Auli. 2019.
\newblock Fairseq: A fast, extensible toolkit for sequence modeling.
\newblock \emph{NAACL HLT 2019}, page~48.

\bibitem[{Papineni et~al.(2002)Papineni, Roukos, Ward, and
  Zhu}]{papineni2002bleu}
Kishore Papineni, Salim Roukos, Todd Ward, and Wei-Jing Zhu. 2002.
\newblock Bleu: a method for automatic evaluation of machine translation.
\newblock In \emph{Proceedings of the 40th annual meeting on association for
  computational linguistics}, pages 311--318. Association for Computational
  Linguistics.

\bibitem[{Pereyra et~al.(2017)Pereyra, Tucker, Chorowski, Kaiser, and
  Hinton}]{pereyra2017regularizing}
Gabriel Pereyra, George Tucker, Jan Chorowski, Lukasz Kaiser, and Geoffrey
  Hinton. 2017.
\newblock Regularizing neural networks by penalizing confident output
  distributions.

\bibitem[{Rajpurkar et~al.(2018)Rajpurkar, Jia, and Liang}]{rajpurkar2018know}
Pranav Rajpurkar, Robin Jia, and Percy Liang. 2018.
\newblock Know what you don’t know: Unanswerable questions for squad.
\newblock In \emph{Proceedings of the 56th Annual Meeting of the Association
  for Computational Linguistics (Volume 2: Short Papers)}, pages 784--789.

\bibitem[{Scialom et~al.(2019)Scialom, Lamprier, Piwowarski, and
  Staiano}]{scialom2019answers}
Thomas Scialom, Sylvain Lamprier, Benjamin Piwowarski, and Jacopo Staiano.
  2019.
\newblock Answers unite! unsupervised metrics for reinforced summarization
  models.
\newblock In \emph{Proceedings of the 2019 Conference on Empirical Methods in
  Natural Language Processing and the 9th International Joint Conference on
  Natural Language Processing (EMNLP-IJCNLP)}, pages 3237--3247.

\bibitem[{Song et~al.(2019)Song, Tan, Qin, Lu, and Liu}]{song2019mass}
Kaitao Song, Xu~Tan, Tao Qin, Jianfeng Lu, and Tie-Yan Liu. 2019.
\newblock Mass: Masked sequence to sequence pre-training for language
  generation.
\newblock In \emph{International Conference on Machine Learning}, pages
  5926--5936.

\bibitem[{Trischler et~al.(2017)Trischler, Wang, Yuan, Harris, Sordoni,
  Bachman, and Suleman}]{trischler2017newsqa}
Adam Trischler, Tong Wang, Xingdi Yuan, Justin Harris, Alessandro Sordoni,
  Philip Bachman, and Kaheer Suleman. 2017.
\newblock Newsqa: A machine comprehension dataset.
\newblock In \emph{Proceedings of the 2nd Workshop on Representation Learning
  for NLP}, pages 191--200.

\bibitem[{Welleck et~al.(2019)Welleck, Weston, Szlam, and
  Cho}]{welleck2018dialogue}
Sean Welleck, Jason Weston, Arthur Szlam, and Kyunghyun Cho. 2019.
\newblock \href {https://doi.org/10.18653/v1/P19-1363} {Dialogue natural
  language inference}.
\newblock In \emph{Proceedings of the 57th Annual Meeting of the Association
  for Computational Linguistics}, pages 3731--3741, Florence, Italy.
  Association for Computational Linguistics.

\bibitem[{Williams et~al.(2018)Williams, Nangia, and
  Bowman}]{williams2018broad}
Adina Williams, Nikita Nangia, and Samuel Bowman. 2018.
\newblock A broad-coverage challenge corpus for sentence understanding through
  inference.
\newblock In \emph{Proceedings of the 2018 Conference of the North American
  Chapter of the Association for Computational Linguistics: Human Language
  Technologies, Volume 1 (Long Papers)}, pages 1112--1122.

\bibitem[{Wolf et~al.(2019)Wolf, Debut, Sanh, Chaumond, Delangue, Moi, Cistac,
  Rault, Louf, Funtowicz et~al.}]{wolf2019transformers}
Thomas Wolf, Lysandre Debut, Victor Sanh, Julien Chaumond, Clement Delangue,
  Anthony Moi, Pierric Cistac, Tim Rault, R{\'e}mi Louf, Morgan Funtowicz,
  et~al. 2019.
\newblock Transformers: State-of-the-art natural language processing.
\newblock \emph{arXiv preprint 1910.03771}.

\bibitem[{Zhang et~al.(2019)Zhang, Kishore, Wu, Weinberger, and
  Artzi}]{zhang2019bertscore}
Tianyi Zhang, Varsha Kishore, Felix Wu, Kilian~Q Weinberger, and Yoav Artzi.
  2019.
\newblock Bertscore: Evaluating text generation with bert.
\newblock \emph{arXiv preprint 1904.09675}.

\end{thebibliography}
